\documentclass[11pt,a4paper]{article}
\usepackage{iclr2021_conference,times}

\usepackage{amsmath,amsfonts,bm}

\def\eqref#1{equation~\ref{#1}}

\def\1{\bm{1}}

\DeclareMathAlphabet{\mathsfit}{\encodingdefault}{\sfdefault}{m}{sl}
\SetMathAlphabet{\mathsfit}{bold}{\encodingdefault}{\sfdefault}{bx}{n}

\usepackage[colorlinks = true, linkcolor = blue, urlcolor  = blue, allcolors=blue]{hyperref}
\usepackage{url}
\usepackage{caption}
\usepackage{subcaption}
\usepackage[legalpaper, portrait, margin=3cm]{geometry}

\usepackage{floatrow}
\newfloatcommand{capbtabbox}{table}[][\FBwidth]

\usepackage{blindtext}

\usepackage{times}
\usepackage{latexsym}

\newcommand{\Enc}{\mathrm{Enc }}
\newcommand{\Dec}{\mathrm{Dec}}
\usepackage{booktabs}
\usepackage{microtype}
\usepackage{graphicx}
\newcommand*{\footcite}[1]{\footnote{\citep{#1}}}
\usepackage{tabularx}
\usepackage{multirow}
\usepackage{booktabs}
\usepackage{amssymb}
\usepackage{tablefootnote}[2014/01/26]

\newcommand{\mcL}{\mathcal{L}}

\title{Pre-trained Summarization Distillation}
\author{Sam Shleifer \thanks{Ask questions  \href{https://discuss.huggingface.co/t/seq2seq-distillation-methodology-questions/1270}{here}, or start your own thread and tag @sshleifer. We are grateful to Stas Bekman, Zoe Shleifer, Patil Suraj and Victor Sanh for comments.}\\
  Hugging Face \\
  \texttt{sam@huggingface.co}  \\\And
  Alexander M. Rush \\
  Hugging Face and Cornell University \\
  \texttt{sasha@huggingface.co} \\}

\date{}

\iclrfinalcopy 

\begin{document}
\maketitle

\begin{abstract}

Recent state-of-the-art approaches to summarization utilize large pre-trained Transformer models. Distilling these models to smaller student models has become critically important for practical use; however there are many different distillation methods proposed by the NLP literature.
Recent work on distilling BERT for classification and regression tasks shows strong performance using direct knowledge distillation. Alternatively, machine translation practitioners distill using pseudo-labeling, where a small model is trained on the translations of a larger model. A third, simpler approach is to ``shrink and fine-tune'' (SFT), which avoids any explicit distillation by copying parameters to a smaller student model and then fine-tuning.
We compare these three approaches for distillation of Pegasus and BART, the current and former state of the art, pre-trained summarization models, and find that SFT outperforms knowledge distillation and pseudo-labeling on the CNN/DailyMail dataset, but under-performs pseudo-labeling on the more abstractive XSUM dataset. PyTorch Code and checkpoints of different sizes are available \href{http://tiny.cc/4iy0tz}{through Hugging Face transformers}.
\end{abstract}

\section{Introduction}

Pre-trained transformer models continue to grow in size \citep{gpt3}, motivating researchers to try to compress large pre-trained checkpoints into smaller, faster versions that retain strong performance.

Recently, researchers have developed promising methods for utilizing pre-trained models for sequence-to-sequence (``Seq2Seq")  language generation tasks, showing particularly large improvements in performance on summarization.
BART \citep{lewis2019bart}, a Seq2Seq transformer \citep{Vaswani2017AttentionIA}  recently achieved state of the art performance on the Extreme Summarization (``XSUM") and CNN/Dailymail(``CNN") summarization datasets \citep{Narayan2018DontGM, DBLP:journals/corr/SeeLM17}, with particularly large improvements on XSUM. A few months later, Pegasus \citep{zhang2019pegasus} achieved further performance improvements by replacing BART's more general pre-training objective with a pre-training objective specifically tailored to abstractive text summarization and a 25\% larger model.
\begin{figure}[!h]
    \centering
    \hspace*{-0.7cm}\includegraphics[scale=0.2]{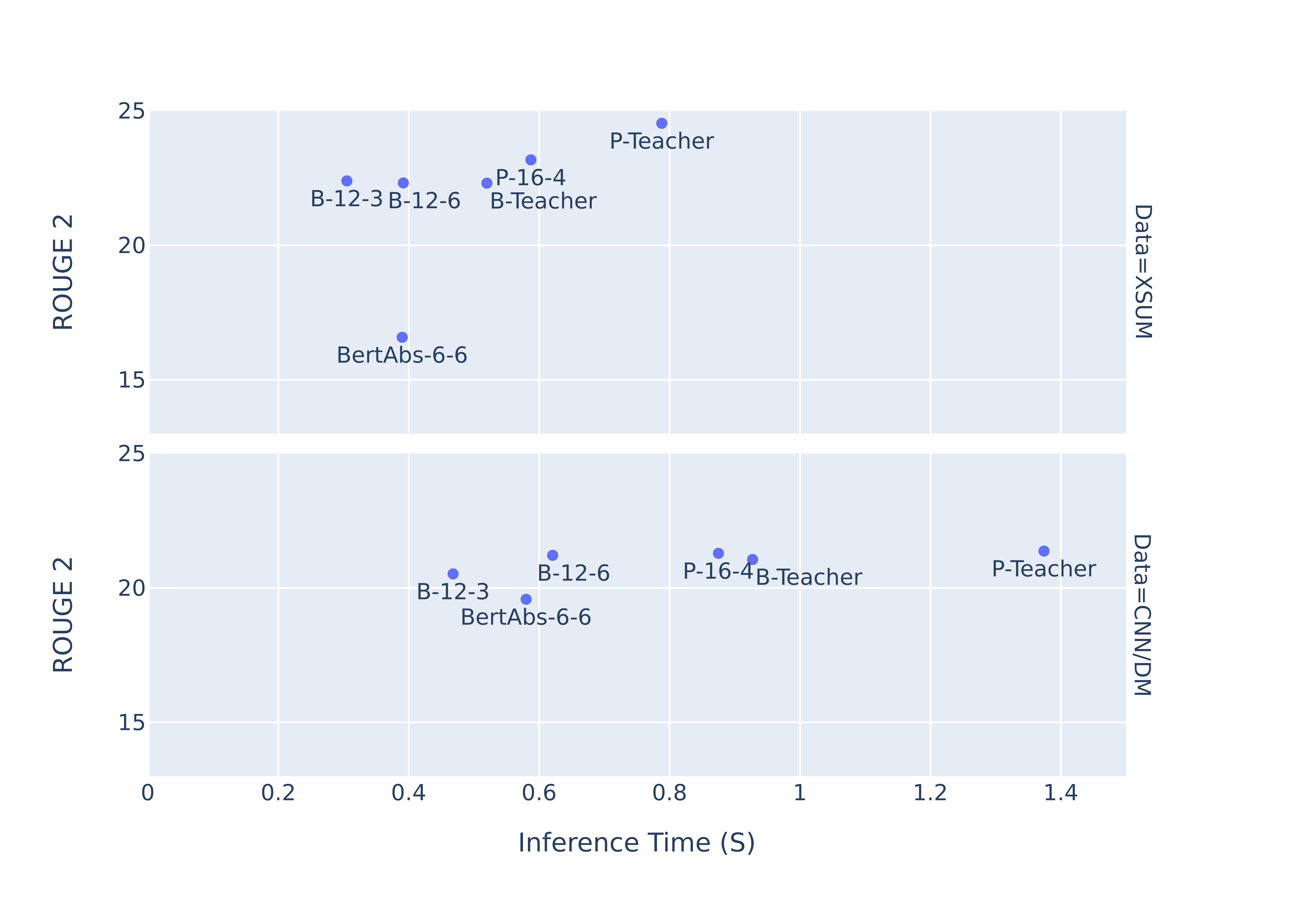}
    \caption{The best distilled checkpoint from Pegasus (P) and Bart (B) for XSUM and CNN at different sizes. In three out of four settings we are able to distill a student model to the same Rouge-2 score as the teacher with at least a 90\% speedup.}
    \label{square}
\end{figure}

In parallel to the progress on summarization, DistilBERT, BERT of Theseus, TinyBERT, MobileBERT, and MiniLM showed that BERT, a large pre-trained transformer, can be shrunk substantially without much performance degradation on the GLUE suite of non-generative tasks using direct Knowledge Distillation (``KD") \citep{sanh2019distilbert, jiao2019tinybert, mobilebert, wang2020minilm, Devlin2018BERTPO, GLUE, Hinton2015DistillingTK}.
On the other hand, past work in machine translation suggests that Seq2Seq models should be compressed with pseudo-labeling (``PL") \citep{Kim_2016}. PL approaches run beam search with the teacher model on the whole training dataset, then retrain a smaller student model from scratch on those translations. Since BART and Pegasus are both pre-trained, like BERT, and Seq2Seq, like the translation models, it is not clear whether Knowledge Distillation or pseudo-labeling is the best approach.
Other approaches are possible as well. Several works suggest that subsets of trained teacher models can be extracted directly \citep{sanh2019distilbert, xu2020bertoftheseus, layerdrop}. We therefore propose a ``shrink and fine-tune" (``SFT") approach that extracts a student model from the maximally spaced layers of a fine-tuned teacher. Since transformer layers are stacked using residual connections, we hypothesize that removing full layers has a minimal impact on summarization performance. This shrunken student model is then used to re-run the original fine-tuning procedure without modification.

We test all three methods on the CNN and XSUM datasets. On CNN, SFT outperforms the more expensive methods. For both BART and Pegasus, SFT produces distilled models that are 75\% faster than their teacher with minimal loss in performance. On the more abstractive XSUM task, KD and PL can generate significant improvements over SFT. For BART, we use KD \footnote{More specifically, we use extensions of KD proposed in \citet{jiao2019tinybert}, and explained in Section \ref{approach}.} to match teacher performance. For Pegasus, no technique matches teacher performance, but PL comes closest. As shown in Figure \ref{square}, we manage to find an approach that generates the best available model at its computational budget for each task and teacher model. In the BART case, we generate many such models of various sizes.

The paper is organized as follows: Section \ref{rw} discusses related work in further detail. Section \ref{approach} describes the specifics of our implementation of the three families of techniques. Section \ref{experiments} describes summarization speed and quality for various teachers, datasets, student sizes, and distillation methods. Sections \ref{pseudo} and \ref{intermediate} describe extensions of pseudo-labeling and knowledge distillation which can further improve performance on the XSUM task.

\section{Related Work}
\label{rw}
\begin{table}
\centering
\begin{tabular}{|c||ccc||cccc|}
\hline
\multirow{2}{*}{Method} &

\multicolumn{3}{c||}{Setup} &
\multicolumn{4}{c|}{Knowledge Transfer} \\
               & Task & Pre-Train Teacher & Pre-Train Student & Init.   & Logits  & Hidden & Gens. \\ \hline
DistilBERT$\heartsuit$     & GLUE & \checkmark               & \checkmark             & \checkmark   & \checkmark   &        &      \\
TinyBERT$\clubsuit$      & GLUE & \checkmark               & \checkmark             &  & \checkmark   & \checkmark         &      \\
BERT-of-Theseus$\diamondsuit$ & GLUE & \checkmark               &            & \checkmark   &  &        &      \\
Seq-Level KD$\spadesuit$  & MT   &              &            &  &  &        & \checkmark       \\
\hline
KD $\dagger$              & Summ. & \checkmark               &            & \checkmark   & \checkmark   & \checkmark         &      \\
Pseudo-Labels $\dagger$   & Summ. & \checkmark               &            & \checkmark   &  &        & \checkmark       \\
SFT $\dagger$            & Summ. & \checkmark               &            & \checkmark   &  &        &      \\ \hline
\end{tabular}
\caption{A comparison of the setting studied and knowledge transfer techniques employed by different transformer distillation methods. $\dagger$ indicates our implementation.  \textsc{Init}: Are weights copied from teacher to student? \textsc{Pre-train Student}: must the student be pre-trained? \textsc{Logits}: does the student learn from the teacher's logits? \textsc{Hidden}: does the student learn from the teacher's hidden states?  \textsc{Gens}: Does the student learn from the teacher's generations?  $\heartsuit$: \cite{sanh2019distilbert},   $\clubsuit$: \cite{jiao2019tinybert},   $\diamondsuit$:  \cite{xu2020bertoftheseus}, $\spadesuit$: \cite{Kim_2016}.}
\label{method_delta}
\end{table}

Knowledge distillation  is a compression technique where a smaller student model is trained to reproduce the logits of a larger teacher, rather than simply minimize the cross-entropy between the model's predicted distribution and the training labels \citep{Bucila2006ModelC, Hinton2015DistillingTK}. In a language modeling context, this allows the student model to learn a full distribution of possible next words in a given context, rather than just the next word in the training data.

Recent research on KD for pre-trained models has overwhelmingly focused on distilling BERT to perform well on GLUE tasks, rather than tasks that require text generation. \citet{sanh2019distilbert} use a weighted average of KD loss and the traditional cross entropy data loss to train DistilBERT, a 6 layer distilled version of BERT, that is 60\% faster on CPU and 50\% faster on GPU.
DistilBERT intializes student models by copying alternating layers. an idea we extend -- in all of our experiments, we initialize students by copying maximally spaced layers.
In TinyBERT, \citet{jiao2019tinybert} add terms to the KD loss function which enforce student/teacher alignment at intermediate levels and improve performance. \footnote{We refer to both of these formulations as KD. DistillBERT can be described as the tinyBERT variant with a zero coefficient on all terms besides the logits loss.}  
Bert-of-Theseus~\citep{xu2020bertoftheseus} randomly replaces multiple teacher layers with a single student layer during fine-tuning with probability $r$, such that each student layer learns to replicate 2 teacher layers. LayerDrop, a related technique, drops random parts of the teacher model during one long training run, allowing a smaller student model to be extracted at inference time \citep{layerdrop}.
Distillation for Seq2Seq models has primarily used pseudo-labeling and produces strong results on machine translation, as shown in \citet{deepShallow}, \citet{junczys-dowmunt-2019-microsoft}, and \citet{sun-etal-2019-baidu}.  Their approach consists of re-generating a new distilled dataset containing original source documents with pseudo-labels. The pseudo-labels are summaries generated by the teacher using beam search. After the long dataset generation process, they train a smaller student model on the ``distilled" dataset. \citet{Kim_2016} call this type ``Sequence-level Knowledge Distillation" in contrast to ``Word-Level Knowledge Distillation", where knowledge is transferred through logits.

Recent work from \citet{liu2020noisy} presents a new method to further improve fine-tuned summarization models by fine-tuning them on their own logits with added noise. Like quantization \citep{jacob2017quantization}, this method could be used before or after the other methods in this work.

Table \ref{method_delta} compares the attributes of these methods to our three approaches. Like Theseus, our experiments do not re-run pre-training. 
SFT is most similar to BERT-of-Theseus, and can even be described as running the Theseus procedure with $r$ fixed at 100\%, thereby saving computation. Our KD implementation is most similar to TinyBERT, and Pseudo-labels is most similar to Sequence Level Knowledge Distillation.

\section{Background and Methods}
\label{approach}

Assume we have a source document $x_{1} \ldots x_{M}$ and target document $y_{1} \ldots y_{N}$ in the standard sequence-to-sequence setting. A Seq2Seq transformer is composed of a transformer-based encoder ($\Enc$) and decoder ($\Dec$). $\Enc$ is trained to map $x$ to contextual embeddings and $\Dec$ to map those contextual embeddings and the previously decoded words to a probability distribution for the next word, $p(y_{t+1} | y_{1:t}, x)$.

Pre-trained Seq2Seq models such as BART and Pegasus learn parameters which are subsequently fine-tuned on Seq2Seq tasks like summarization. BART is pre-trained to reconstruct corrupted documents. Source documents $x$ are corrupted versions of original target documents $y$, e.g. spans of words are masked and sentences are shuffled. Pegasus is pre-trained to generate the most important sentences extracted from an unlabeled document ($y$ is the important sentences and $x$ is the original documented with those sentences removed).

To fine-tune the models, we assume a dataset where each example $(x,y)$ is a (document, summary) pair.
In the Seq2Seq fine-tuning setting, we train the student model using the standard cross entropy loss:
\begin{equation}
\mcL_{\text{Data}} = -\sum_{t=1}^{T} \log p(y_{t+1} | y_{1:t}, x)
\end{equation}
where $T$ in the target sequence length $p$ is the model's predicted probability for the correct word. Our distillation experiments start with a teacher model fine-tuned in this manner.

\subsection{Distillation}

We consider different approaches for compressing these models through distillation. All settings assume that we are learning a student model from a larger teacher.
Define the notation $\Dec^L$ to represent a decoder with $L$ Transformer layers (and similarly for $\Enc$). Assuming we have a large pre-trained teacher model with decoder $\Dec^L$, we are interested in compressing it to a smaller student model $\Dec^{L'}$. In most experiments, we do not compress the teacher's encoder.

\paragraph{Shrink and Fine-Tune}
Our most basic method SFT simply \textit{shrinks} the teacher model to student size and re-fine-tunes this student model. Here  each $l \in L'$ is copied fully from $L$; students are initialized by copying full maximally spaced decoder layers from teacher to student. For example, when creating a BART student with 3 decoder layers from the 12 encoder layer 12 decoder layer teacher, we copy the teacher's full $\Enc^L$ and decoder layers 0, 6, and 11 to the student. When deciding which layers to copy, we break ties arbitrarily; copying layers 0, 5, and 11 might work just as well. When copy only 1 decoder layer, we copy layer 0. We found this to work better than copying layer 11. The impact of initialization on performance is measured experimentally in Section \ref{init_exp_section}. After initialization, the student model continues to fine-tune on the summarization dataset, with the objective of minimizing $\mcL_{\text{Data}}$. As the initialization approach is simple and effective, it is used to initialize student models for both other methods.

\paragraph{Pseudo-labels} In the pseudo-label setting, we replace the ground truth target documents $Y$ with $\hat{Y}$, the teacher's generations for the source documents $X$, computed with beam search.
\begin{equation}
\mcL_{\text{Pseudo}} = -\sum_{t=1}^{T} \log p(\hat{y}_{t+1} | \hat{y}_{1:t}, x)
\end{equation}
After this procedure the student model is fine-tuned only on this new pseudo-labeled data.

\paragraph{Direct Knowledge Distillation (KD)} In the KD setting, even more information is transferred from teacher to student, by encouraging the student to match the teacher's full probability distribution over possible next words at each position, by minimizing KL-Divergence\citep{kullback1951information, sanh2019distilbert}:
\begin{equation}
\mcL_{\text{Logits}} = \sum_{t=1}^{T} KL(Q_{t+1}, P_{t+1}),
\end{equation}
where $Q_{t+1}$ and $P_{t+1}$ are teacher and student probability distributions over each next possible word at position $t+1$, and $KL$ is the KL-Divergence.\footnote{KL-Divergence is implemented in PyTorch \citep{paszke2019pytorch} and explained well on \href{http://tiny.cc/wiki_kldiv}{Wikipedia}.}
Since we use layer based compression, student and teacher layers output the same shape, and we can add another term to the loss function that encourages students to match teacher hidden states. 
\begin{equation}
\label{eq:hid_loss}
\mathcal{L}_{\text{Hid}} = \sum_{t=1}^{T} \sum_{l=1}^{L'} \texttt{MSE}(\bm{H}_{l}^{S}, \bm{H}_{\phi(l)}^{T}) 
\end{equation}
Here, $\texttt{MSE}$ stands for mean squared error, $\bm{H}_{l}^{S}$ retrieves the hidden state returned by student layer $l$, and $\phi(l)$ maps student layer $l$ to the teacher layer whose output we would like them to emulate. $\bm{H}_{\phi(l)}^{T}$, therefore, is the output of a teacher layer. \footnote{For a more detailed description of $\mcL_{\text{Hidn}}$, read the Methods Section of the \href{https://arxiv.org/pdf/1909.10351.pdf}{TinyBert paper}. Our approach is inspired by theirs, but we do not use per-layer weights, per-layer learning rates, embedding loss, or attention loss.}
For example, when creating a BART student with 3 decoder layers, we copy the full teacher encoder and decoder layers 0, 6, and 11 to the student. We then choose pairings in $\phi$ such that each student decoder layer is taught to behave like 4 decoder layers. Student layer 0's hidden state is paired o teacher layer 3, 1 to 7, and 11 to 11 ($\phi = [3,7,11]$). The student layers are therefore trained to perform the work of teacher layers 0-3, 4-7 and 8-11 respectively. \footnote{A complete list of the $\phi$ mappings we used can be found \href{https://tinyurl.com/y2qlacbm}{here}}

Our final KD formulation is a weighted average: 
\begin{equation}
\mcL_{\text{KD}} = \alpha_{logits} \mcL_{\text{Logits}} + \alpha_{data} \mcL_{\text{Data}} + \alpha_{Hidn} \mcL_{\text{Hidn}}
\end{equation}
We set $\alpha_{logits}=0.8$ and $\alpha_{data}=1$ following \cite{sanh2019distilbert}, and found $\alpha_{Hidn}=3$ to perform best out of $[1, 3, 10, 100]$ for BART on the XSUM development set.

\paragraph{Training Time Comparison} Table \ref{tab:wrap_tech} compares the training time of these three approaches. Whereas SFT simply requires fine-tuning a small model, computing $\mathcal{L}_{\text{KD}}$ requires teacher logits; for each training example, we must run the large teacher model forwards as well as the student model forwards and backwards. Similarly, $\mcL_{\text{Pseudo}}$ requires $\hat{Y}$, which is computed by running beam search with the teacher on the full training dataset. This large preprocessing cost can dwarf the cost of fine-tuning the student model on the pseudo-labels, as shown in Table \ref{tab:wrap_tech}, where 16.5 of the 19 GPU hour cost of producing a student model is spent generating the pseudo-labels. After initialization, SFT does not use the teacher model, and is therefore much cheaper.


\begin{table}{}
\centering
\begin{tabular}{@{}cccc@{}}
\toprule
Technique         & Extra Supervision       & Cost & Loss\\ \midrule
SFT               &                         & 2.5 & $\mcL_{\text{Data}}$            \\
PseudoLabeling    & T's Generations         & 19 & $\mcL_{\text{Pseudo}}$           \\ 
KD  & T's Hidden States, Logits & 14  & $\mcL_{\text{KD}}$         \\ \bottomrule
\end{tabular}
\caption{Training time of different distillation approaches. Cost is an estimate of how many hours were required to run the technique for the CNN dataset with BART as a teacher and the 12 encoder layer, 6 decoder layer student on a Titan RTX 2080 GPU.}
\label{tab:wrap_tech}
\end{table}

\section{Experimental Setup}
\begin{table}
\centering
\begin{tabular}{@{}ccccc@{}}
\toprule
Data & \# Train & Avg. Source Words & Avg. Target Words & Size (MB) \\ \midrule
CNN   & 262,567  & 756               & 56                & 1,331           \\
XSUM  & 204,017  & 358               & 21                & 501             \\
\midrule
EN-RO & 610,319  & 23                & 23                & 178             \\
\bottomrule
\end{tabular}
\caption{Dataset Statistics.}
\label{tab:ds_stats}
\end{table}
We experiment with both the CNN and XSUM abstractive summarization datasets, both of which are based on english language news articles.
The CNN summaries are roughly 3 sentences long, and tend to be similar to text from the beginning of the document.
The XSUM summaries are the first sentence of a BBC news article, which is then removed from the article, so are both shorter and more abstractive than CNN summaries. The original BART model's improvement over its predecessors was much more significant (roughly 6 ROUGE-2 points) on the more abstractive XSUM dataset than on the CNN dataset (1.5 points). Table \ref{tab:ds_stats} shows dataset statistics. 

\paragraph{Generation and Evaluation} We run beam search on the distilled models and measure summary quality using the Rouge implementation from the \texttt{rouge\_scorer} python package. ROUGE scores for teachers and students can be found in Tables \ref{XSUM_rouge} and \ref{cnn_rouge}. Unlike the BART paper, we do not tokenize or otherwise preprocess summaries before scoring, leading to slightly lower scores. For inference speed comparison, we measure Summaries per Second using a batch size of 32 and mixed precision for BART on 1 GPU. Mixed precision overflows for Pegasus, so we use full precision.  

\paragraph{Translation Experiments}
Translation experiments are included for comparison. These use the  English-Romanian WMT 2016 English-Romanian Dataset (``EN-RO'') \citep{bojar-etal-2016-findings}, and two teachers. ``mBART'' is pre-trained on many languages and then fine-tuned on bilingual data. \citep{MBART},  ``Marian'' is trained from scratch on bilingual data \citep{Marian}. These experiments are evaluated using BLEU with no post-processing \citep{papineni2002bleu}.

\paragraph{Training}
We stop training at whichever point comes first: the end of epoch 5 or the validation score not increasing for four consecutive evaluations (a full epoch). We measure ``training cost" as the amount of time training takes on one Nvidia-RTX-2080 GPU + the cost of generating pseudo-labels, if applicable.

In experiments with a full sized (completely copied) encoder, we freeze its parameters during training. Initial experiments suggested that this did not impact performance but made fine-tuning faster by a factor of 5.\footnote{For KD, if the encoder is the same for teacher and student, it only needs to be run once. Back propagation is also much cheaper, as it can stop at the end of the encoder.}
We also freeze the positional and token embeddings. 

\paragraph{Effort} 
We did not spend equal resources on all datasets and models, as shown in Table \ref{effort}.
\begin{table}
\centering
\begin{tabular}{@{}cccccc@{}}
\toprule
Teacher & Dataset & \# GPU Hours & \% GPU Hours & \# Experiments & \% Experiments \\ \midrule
BART    & XSUM    & 787          & 30\%         & 102            & 36\%           \\
BART    & CNN  & 365          & 14\%         & 59             & 21\%           \\
mBART   & EN-RO   & 332          & 13\%         & 48             & 17\%           \\
Pegasus & XSUM    & 766          & 29\%         & 42             & 15\%           \\
Marian  & EN-RO   & 185          & 7\%          & 26             & 9\%            \\
Pegasus & CNN  & 196          & 7\%          & 10             & 3\%            \\ \midrule
        & TOTALS  & 2,631         &              & 287            &                \\ \bottomrule
\end{tabular}
\caption{Effort calculations. Each row represents the resources spent attempting to distill a teacher to a smaller student model on a given dataset. Experiments were only counted if they lasted 15 minutes or more. \% columns divide \# columns by their sum.}
\label{effort}
\end{table}
In particular, we ran fewer CNN experiments because SFT worked well in that case, and fewer Pegasus experiments because Pegasus takes longer to train. Many of the BART experiments on XSUM tested variants and hyperparameters for KD, which has yet to work well for Pegasus. If we had run 60 more Pegasus experiments on XSUM data, we might have found something that works better.

\paragraph{Model Notation}
We use shorthand notation to describe student models generated with our initialization procedure. For example, \texttt{dBART-12-3} is a student model extracted from BART with (all) 12 encoder layers and 3 decoder layers. Similarly, all ``Size" columns in tables use the Encoder Layers-Decoder Layers convention.

\section{Results} 
\label{experiments}

Table \ref{by_approach} shows the performance of 3 different approaches for different tasks, teachers and students. No approach dominates the others across all datasets.
On CNN, SFT works best for both teachers. On XSUM, BART performs best with KD, while Pegasus performs best with PL. ROUGE-1 and ROUGE-L scores follow a similar pattern to ROUGE-2 in Table \ref{by_approach} on both summarization datasets. We additionally include translation experiments for comparison. On the English-Romanian translation dataset, PL works best for both teacher models.  

Tables \ref{XSUM_rouge} and \ref{cnn_rouge} show scores and inference times for many different student models on XSUM and CNN, respectively. These tables show the best student of a given size, regardless of distillation method. In 3 out of 4 contexts, distillation leads to relatively minor performance losses and significant speedups. On XSUM, both the 12-3 and 12-6 sized BART students outperform the teacher model at 93\% and 43\% speedups, whereas the Pegasus student falls more than a full ROUGE-2 point below the teacher model. On CNN, the 12-6 sized BART student outperforms the teacher, and the Pegasus teacher is close.

Note that Table \ref{XSUM_rouge} shows a higher score for the BART/XSUM 12-3 student than Table \ref{by_approach} shows. The stronger student was trained on pseudo-labels generated by Pegasus. The result is not included in Table \ref{by_approach}'s PL column, which shows results for student models trained on pseudo-labels generated by their teacher. We discuss this further in Section \ref{pseudo}.

\begin{table}
\centering
\begin{tabular}{@{}cccccccccc@{}}
\toprule
Teacher & Size & Data & Teacher & \multicolumn{2}{c}{SFT} & \multicolumn{2}{c}{KD} & \multicolumn{2}{c}{Pseudo} \\ 
        &      &        & Score  & Score   & Cost & Score & Cost & Score & Cost \\ \midrule
BART $\dagger$    & 12-3 & XSUM   & 22.29  & 21.08   & 2.5  & \textbf{21.63 }& 6    & 21.38 & 15   \\
Pegasus & 16-4 & XSUM   & 24.56  & 22.64   & 13   & 21.92 & 22   & \textbf{23.18} & 34   \\
BART    & 12-6 & CNN & 21.06  & \textbf{21.21}   & 2    & 20.95 & 14   & 19.93 & 19.5   \\
Pegasus    & 16-4 & CNN & 21.37  & \textbf{21.29}   & 31 &   - & - & 20.1   & 48    \\
\midrule
Marian  & 6-3  & EN-RO  & 27.69  & 25.91   & 4    & 24.96 & 4    & \textbf{26.85} & 28   \\
mBART   & 12-3 & EN-RO  & 26.457 & 25.6083 & 16   & 25.87 & 24   & \textbf{26.09} & 50   \\ \bottomrule
\end{tabular}
\caption{Main results. Score is Rouge-2 for the 2 summarization datasets (first 4 rows), and BLEU for the bottom two rows. Cost measures the GPU hours required to run the approach end to end, which, in the case of Pseudo-labeling, requires running beam search on the full training set. The highest scoring distillation technique is in bold.
}
\label{by_approach}
\end{table}

\begin{table}[!h]
\centering
\begin{tabular}{@{}lllcccc@{}}
\toprule
Teacher     & Student          & MM Params &  Time (MS) & Speedup & Rouge-2 & Rouge-L \\
\midrule
BART    & 12-1             & 222       & 743                 & 2.35    & 17.98   & 33.31   \\
        & 12-3             & 255       & 905                 & 1.93    &\textbf{ 22.40}   & 37.30   \\
        & 6-6              & 230       & 1179                & 1.48    & 21.17   & 36.21   \\
        & 9-6              & 268       & 1184                & 1.47    & 22.08   & 37.24   \\
        & 12-6             & 306       & 1221                & 1.43    & 22.32   &\textbf{ 37.39}   \\
        & Baseline (12-12) & 406       & 1743                & 1.00    & 22.29   & 37.20   \\
\midrule
Pegasus &        16-4             & 369       & 2038                & 2.40    & 23.18   & 38.13   \\
       & 16-8             & 435       & 2515                & 1.94    & 23.25   & 38.03   \\
         & Baseline (16-16) & 570       & 4890                &         & \textbf{24.46 }  & \textbf{39.15} \\

 \midrule
 BertABS & Baseline (6-6)   & 110       & 1120                &         & 16.50   & 31.27   \\
 \midrule 
\end{tabular}
\caption{Best XSUM results across all methods. Each sub-table is sorted fastest to slowest by inference time. \texttt{dBART-12-3} and  \texttt{dPegasus-16-4} are trained on Pegasus pseudo-labels. \texttt{dBART-12-6}, \texttt{dBART-6-6}, and \texttt{dBART-9-6} are trained with KD. \texttt{dPegasus-16-8} and \texttt{dBART-12-1} are trained with SFT. For the BART experiments where the encoder is smaller than 12 layers, we do not freeze it during training.}
\label{XSUM_rouge}
\end{table}

\begin{table}[!h]
\centering
\begin{tabular}{@{}lllcccc@{}}
\toprule
Teacher     & Student          & MM Params & Inference Time (MS) & Speedup & Rouge-2 & Rouge-L \\
\midrule
BART        & 12-3             & 255       & 1483                & 1.66    & 20.57   & 40.31   \\
  & 6-6              & 230       & 1684                & 1.46    & 20.17   & 39.55   \\
        & 12-6             & 306       & 1709                & 1.44    &\textbf{ 21.19}   & \textbf{41.01 }  \\
    & Baseline (12-12) & 406       & 2461                & 1.00    & 21.08   & 40.89   \\
\midrule
Pegasus & 16-4             & 369       & 3728                & 2.67    & 21.29   & 40.34   \\
 & Baseline (16-16) & 570       & 9965                &         & \textbf{21.37}   &\textbf{ 41.04 }  \\
\midrule
BertABS & Baseline (6-6)   & 110       & 1582                &         & 19.6    & 39.18     
\\ \bottomrule
\end{tabular}
\caption{Best CNN/Daily Mail Results across all methods, which is always SFT.}
\label{cnn_rouge}
\end{table}

\section{Analysis}





\subsection{How does initialization impact distillation?}
\label{init_exp_section}
\begin{figure}[!h]
    \centering
    \hspace*{-1.5cm}\includegraphics[scale=0.5]{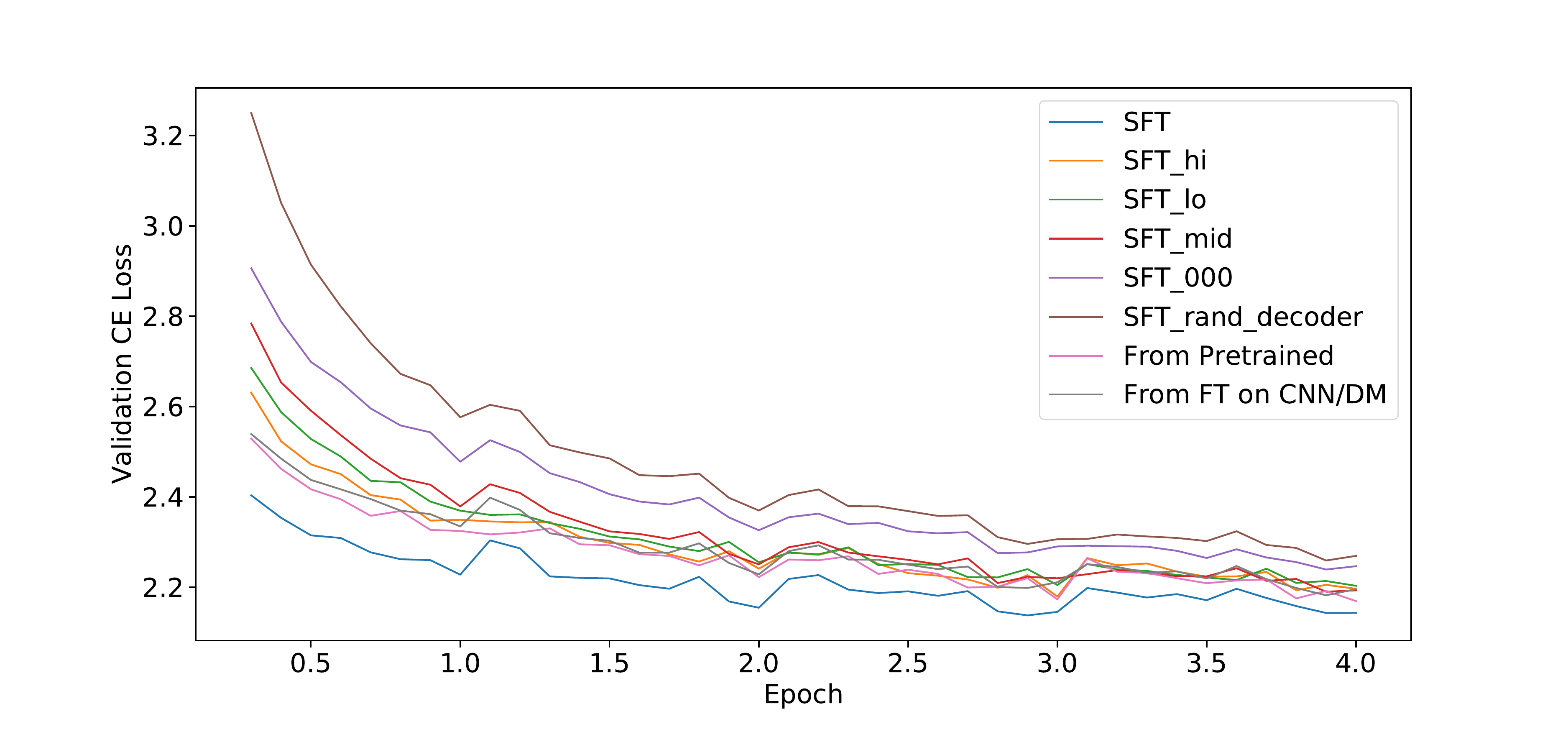}
    \caption{Training curves for different initialization strategies.
    Each line represents one fine-tuning run for a BART student on XSUM using a different initialization strategy. Initialization strategies are described in Table \ref{init_results}.}
    \label{init_experiments}
\end{figure}

In Table \ref{init_results}, we show the validation cross entropy loss of \texttt{dBART-12-3} students trained with the same, frozen encoder, but different decoder layers copied from different sources. The default SFT initialization for 3 layer students, copying layers 0, 6, 11, (the low, blue line in Figure \ref{init_experiments})  converges more quickly and to a better loss than other initialization strategies. We show that this result holds on the CNN and EN-RO datasets in Table \ref{init_cnn}.

\begin{table}[!h]
\centering
\begin{tabular}{@{}lccc@{}}
\toprule
Name                 & Layers Copied & From      & Min Loss \\ \midrule
SFT           & 0,6,11        & XSUM & \textbf{2.14}       \\
SFT hi            & 9,10,11       & XSUM & 2.18          \\
SFT lo            & 0,1,2         & XSUM & 2.20      \\
SFT mid           & 5,6,7         & XSUM & 2.19         \\

SFT 000           & 0,0,0         & XSUM & 2.24         \\
SFT rand decoder & -        & -         & 2.26           \\ 
From Pre-trained    & 0,6,11        & PT   & 2.17       \\
From FT on CNN  & 0,6,11        & CNN  & 2.18        \\
\bottomrule
\end{tabular}
\caption{Loss for different initialization strategies on XSUM. Each row represents one fine-tuning run for a BART student on XSUM using a different initialization strategy (and one line in Figure \ref{init_experiments}.) \textsc{Layers Copied} indicates which decoder layers were copied from the teacher. \textsc{From} indicates the BART model the layers were copied from, where XSUM is the BART teacher fine-tuned on the correct dataset, CNN is the teacher fine-tuned on the wrong dataset, and PT is the pre-trained (but not fine-tuned) BART checkpoint. \textsc{Min Loss} is cross entropy on the XSUM dev set. Validation loss was checked 10 times every epoch.
This table corresponds to the figure above it.
}
\label{init_results}
\end{table}
\begin{table}
\centering
\begin{tabular}{@{}lccc@{}}
\toprule
Name              & Layers Copied & From     & Min Loss        \\ \midrule
SFT             & 0,6,11        & CNN & \textbf{2.02}  \\
From Pre-trained & 0,6,11        & PT  & 2.17                  \\
SFT hi          & 9,10,11       & CNN & 2.31            \\
SFT rand        & -        & -        & 3.91             \\ \midrule
SFT   & 0, 2, 5       & Marian      &\textbf{ 1.66 }          \\
SFT Back  & 3,4,5         & Marian      & 1.70           \\
SFT Front & 0,1,2         & Marian      & 1.88         \\
SFT rand    & -      & -           & 2.2           \\ \bottomrule
\end{tabular}

\caption{Loss for different initialization strategies. See Table~\ref{init_results} for column descriptions. The top half of the table uses BART as a teacher and CNN as a dataset, the bottom half uses the fine-tuned Marian MT model as a teacher and EN-RO as a dataset. In the \textsc{From} column, CNN is BART fine-tuned on CNN, and PT is the pre-trained (but not fine-tuned) BART checkpoint, and Marian is the fine-tuned Marian MT checkpoint, which uses 6 encoder layers and 6 decoder layers.}
\label{init_cnn}
\end{table}

\subsection{When does Pseudo-Labeling help performance?}
\label{pseudo}
Table \ref{pl_variants} shows results from fine-tuning teacher models on combinations of real labels and pseudo-labels.
The Orig and Orig+PL columns show that, for summarization on XSUM, PL can improve over the SFT baseline when the pseudo-labels are added to the original fine-tuning dataset. For translation, (EN-RO), pseudo-labels can simply replace the original training data. For CNN (not shown), PL performs worse than SFT.\footnote{All pseudo-labels are made available for download \href{https://github.com/huggingface/transformers/blob/master/examples/seq2seq/precomputed_pseudo_labels.md}{here}.}
For BART on XSUM, fine-tuning on the original dataset (SFT) generates a student that is 1.2 ROUGE-2 points worse than the teacher, fine-tuning on the original dataset and Pseudo-labels generates a better student, that is only 0.8 points behind the teacher. Adding Pseudo-labels generated by Pegasus, (the Orig+PL+PL column), generates a substantial improvement: the finetuned student is 0.1 points better than the teacher.

For Pegasus on XSUM, however, there is no benefit to adding pseudo-labels generated by BART. Comparing Orig+PL to Orig+PL+PL in Table \ref{pl_variants} Row 2 shows that a student trained on the original data and Pegasus pseudo-labels is 1.2 ROUGE-2 below the teacher, whereas a student trained on the original data, Pegasus pseudo-labels, and BART pseudo-labels is 1.6 ROUGE-2 below the teacher.

The quality of the pseudo-labels may be driving this pattern. If we take the ROUGE-2 of pseudo-labels (against the training set labels) as proxy for their quality, the quality of the Pegasus pseudo-labels is 4 points higher than BART.
Additionally, we did not find that pseudo-labels helped on CNN, where ROUGE scores are lower for both teachers, supporting the quality hypothesis.


\begin{table}
\centering
\begin{tabular}{@{}lllccccc@{}}
\toprule
Teacher & Size & Dataset & Teacher Score & Orig & PL & Orig+PL & Orig+PL+PL$^{*}$ \\ \midrule
BART    & 12-3 & XSUM    & 22.3          & -1.2          & -0.9         & -0.8  & \textbf{+0.1}  \\
Pegasus & 16-4 & XSUM    & 24.5          & -1.9          & -2.2         & \textbf{-1.2  } & -1.6  \\
BART    & 12-3 & CNN  & 21.1          &\textbf{ -1.4 }         & -2.0         & -2.0   & - \\
Pegasus    & 16-4 & CNN  & 21.37          &\textbf{ -0.1}         & -1.4         & -   & - \\ \midrule
Marian  & 6-3  & EN-RO   & 27.7          & -1.8          & \textbf{-0.8}        & -1.8 & -   \\
mBART   & 12-3 & EN-RO   & 26.5          & -0.8          & \textbf{-0.4}         & -0.6 & -   \\ \bottomrule
\end{tabular}
\caption{Pseudo-labeling Strategies. Columns (Orig, PL, Orig+PL, and Orig+PL+PL$^{*}$) report student scores relative to their teacher using (the original training data,  pseudo-labels generated by the Teacher, both, and all pseudo-labels available for a given dataset + the original data).
The score units are ROUGE-2 for the top four rows, BLEU for the two bottom rows, with the score for each student subtracted from the teacher score.  All students are initialized by copying maximally spaced layers from the teacher and trained for 2 epochs.}
\label{pl_variants}
\end{table}

\subsection{Do changes to $\mcL_{kd}$ improve performance?}
\label{intermediate}
Except for BART on XSUM, KD did not generate improvements over SFT, and, as previously discussed, is always more expensive. This was not for lack of effort. Here are few modifications that did not improve performance:

\begin{enumerate}
    \item Removing $\mcL_{\text{Hidn}}$, which encourages student layer $l$ to produce the same hidden state as teacher layer ${\phi{L}}$, hurt performance for BART on XSUM. In the other settings, removing $\mcL_{\text{Hidn}}$ had a negligible affect on performance.
    \item Adding TinyBERT's $\mcL_{\text{Attn}}$, which encourages student layer $l$ to produce the same attention weights as teacher layer $M\_{\phi{L}}$, further slowed training without improving performance. \citep{jiao2019tinybert}
    \item Adding the cosine loss used in DistillBERT to $\mcL_{kd}$ did not impact performance. \citep{sanh2019distilbert}
\end{enumerate}

This suggests that more work is needed for adapting KD approaches that work on BERT to Seq2Seq tasks, and that practitioners should try SFT first, followed by pseudo-labeling.

\subsection{Inference Time Analysis}

To further understand why the 6-6 models ran slower than 12-3 models in Tables \ref{XSUM_rouge} and \ref{cnn_rouge}, we ran a single forward pass on 12,000 different randomly initialized BART configurations in a GPU half-precision environment, and estimated the effects of changing the number of encoder layers, feed forward dimensions, number of decoder layers, and embedding size (width) on inference time with a linear regression. The results suggest that adding a decoder layer would slow down inference by 8\%, while adding an encoder layer would slow down inference by only 4\%. We also observed that changing width or feed forward dimensions had negliglible impact on run time.\footnote{\citet{sanh2019distilbert} found similar results with respect to the BERT architecture; \citep{deepShallow} found similar results for MT.} This difference is exacerbated during beam search, where the decoder is run \texttt{beam\_size} times per example.

\section{Conclusion}
In this paper, we show that for summarization tasks, removing carefully chosen decoder layers from a Seq2Seq transformer and then continuing fine-tuning generates high quality student models quickly, and that in some situations more expensive training techniques with the same initialization strategy can generate additional quality improvements.

Future experiments could (1) evaluate these techniques on other summarization datasets, other tasks, and other teachers, like T5\footcite{raffel2020exploring}. (2) Explore distilling the knowledge in pre-trained, but not fine-tuned, Seq2Seq models. (3) Explore more of the large KD hyper-parameter space. (4) Explore strategies to improve pseudo-label quality. (5) Our experiments target speedups on GPU, but SqueezeBERT \footcite{squeezebert} suggests that reducing the width of each student layer is key to unlocking more efficient CPU inference. \footnote{ \href{https://discuss.huggingface.co/t/seq2seq-distillation-methodology-questions/1270}{Discussion here}}
\newpage

\bibliographystyle{unsrtnat}
\bibliography{my}

\appendix

\end{document}